\documentclass{svproc}
\usepackage{url}

\usepackage{booktabs}
\usepackage{comment}
\usepackage{graphicx}
\usepackage{xcolor}
\usepackage{cite}
\usepackage{hyperref}
\newcommand\sh[1]{\textbf{\textcolor{blue}{SH: #1}}}

\newcommand{\ns}[1]{\textcolor{magenta}{NS: #1}}

\begin{document}
\mainmatter              
\title{Protecting Vulnerable Voices: Synthetic Dataset Generation for Self-Disclosure Detection}
\titlerunning{Synthetic Dataset Generation}  
%
\author{Shalini Jangra\inst{1} \and Suparna De
 \and Nishanth Sastry\and Saeed Fadaei}
\authorrunning{Shalini Jangra et al.} 
%
\tocauthor{Shalini Jangra, Suparna De, Nishnath Sastry, and Saeed Fadaei}
\institute{University of Surrey, UK\\
\email{s.jangra@surrey.ac.uk}, \email{s.de@surrey.ac.uk}, \email{n.sastry@surrey.ac.uk}, \email{s.fadaei@surrey.ac.uk}}

\maketitle              

\begin{abstract}
Social platforms such as Reddit have a network of communities of shared interests, with a prevalence of posts and comments from which one can infer users' Personal Information Identifiers (PIIs). While such self-disclosures can lead to rewarding social interactions, they pose privacy risks and the threat of online harms. Research into the identification and retrieval of such risky self-disclosures of PIIs is hampered by the lack of open-source labeled datasets. Important hindrances to sharing high-quality labelled data include high annotation costs and privacy risks associated with the release of datasets containing self-disclosive text, especially when users include vulnerable populations. 

To foster reproducible research into PII-revealing text detection, we develop a novel methodology to create synthetic equivalents of PII-revealing data that can be safely shared. Our contributions include creating a taxonomy of 19 PII-revealing categories for vulnerable populations and the creation and release of a synthetic PII-labeled multi-text span dataset generated from 3 text generation Large Language Models (LLMs), Llama2-7B, Llama3-8B, and zephyr-7b-beta, with sequential instruction prompting to resemble the original Reddit posts. The utility of our methodology to generate this synthetic dataset is evaluated with three metrics: First, we require \textit{reproducibility equivalence}, i.e., results from training a model on the synthetic data should be comparable to those obtained by training the same models on the original posts. Second, we require that the synthetic data be \textit{unlinkable} to the original users, through common mechanisms such as Google Search. Third, we wish to ensure that the synthetic data be \textit{indistinguishable} from the original, i.e., trained humans should not be able to tell them apart. 
We release our dataset and code at \textcolor{blue}{\url{https://netsys.surrey.ac.uk/datasets/synthetic-self-disclosure/}} to foster reproducible research into PII privacy risks in online social media.
\keywords{Personal Information Identifiers, Synthetic data, Vulnerable Populations, Privacy Leaks, Large Language Models}
\end{abstract}
\section{Introduction}
\label{sec:introduction}
\begin{figure}[t]
  \centering
  \includegraphics[width=.6\textwidth]{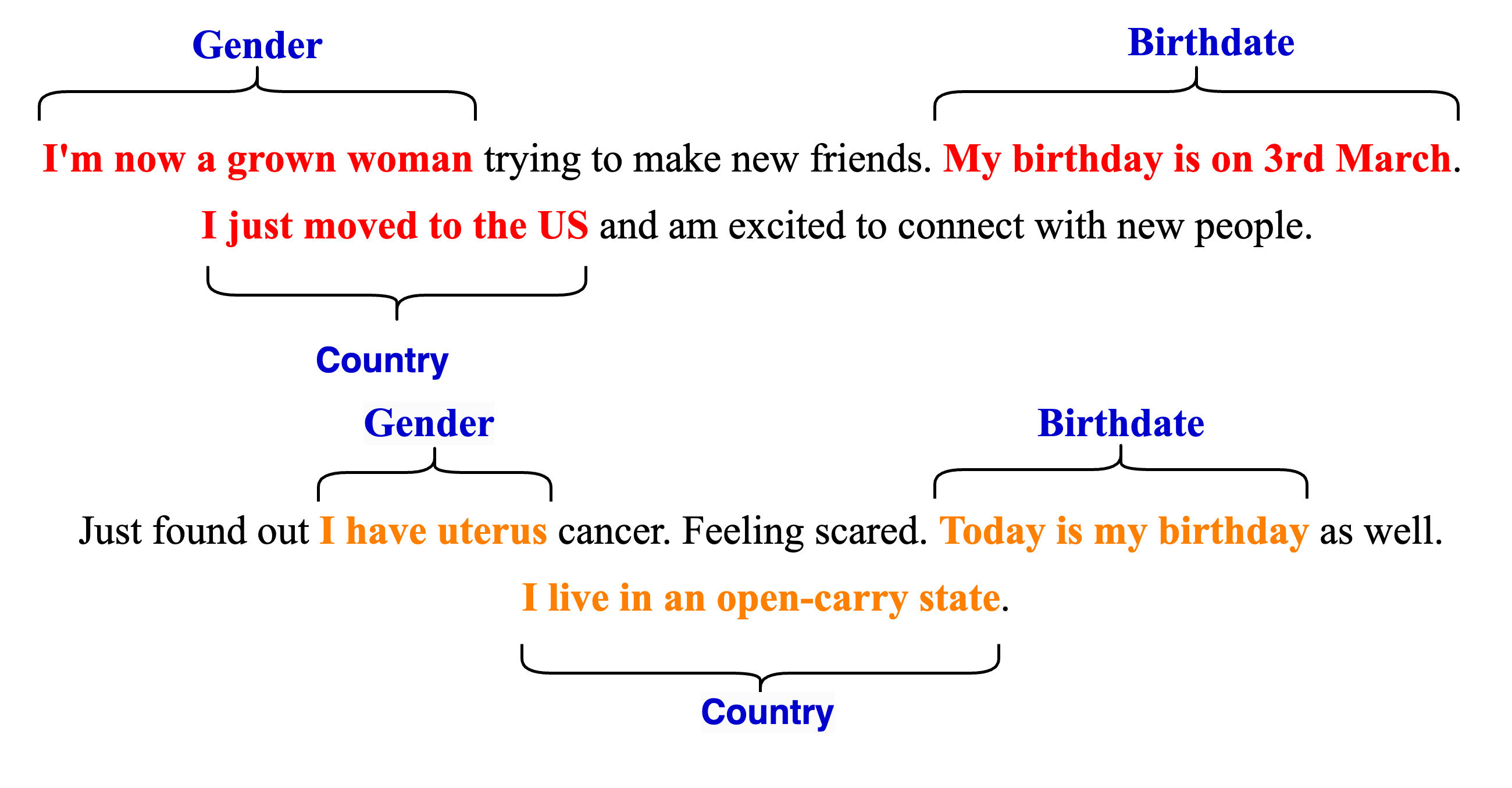}
  \vspace{-0.5cm}
  \caption{Explicit (in \textcolor{red}{red}) and Implicit (in \textcolor{orange}{orange}) PII self-disclosures of gender, birth date and country in Reddit posts}
    \vspace{-0.5cm}
  \label{fig:explicit-implicit}
\end{figure}

Leakage of Personally Identifiable Information (PII) on social media is a common and serious problem: technical affordances such as anonymity, visibility control and editability \cite{corvite2022social} give users opportunities for self-disclosure and support seeking \cite{andalibi2016understanding,mcdonald2022privacy}; activities which may lead them to reveal personal information on social media \cite{mcdonald2022privacy,nicol2022revealing}.
This problem becomes more acute for vulnerable populations who may be targeted after self-disclosure. 
In this work, we consider vulnerable individuals as those who, due to intersecting factors such as race, socioeconomic status, gender or sexual identity, religion, or other marginalized social positions, face an increased risk of privacy violations that may lead to emotional, financial, or physical harm, as characterized in \cite{mcdonald2022privacy}. PII leakage may happen naturally in the course of posts made on social media as demonstrated in \autoref{fig:explicit-implicit}. Such leakage may happen in an \textit{explicit} manner, wherein the text of a post directly reveals personal information, 
or \textit{implicitly}, when PII can be inferred indirectly from a comment that the user makes. For instance, while the explicit mention of `3rd March' for birth-date can be flagged relatively straightforwardly as PII-revealing, talking about `uterus cancer' may indirectly reveal (birth) gender. 

Public datasets have been crucial for reproducible socio-technical research, enabling data-driven solutions to online harms \cite{zubiaga2022editorial}. However, in many cases, the collection and use of real-world datasets raises privacy concerns, as such data may contain potentially sensitive data, even when subjected to deidentification and sampling techniques before release \cite{Narayanan2009, rocher2019}. This is especially true of datasets regarding PII disclosures. 
Even when datasets are based on publicly available data (e.g., public posts on platforms such as X or Reddit), there are ethical challenges about consent~\cite{buck2021didn} and increased visibility~\cite{zimmer2020but}. 

Thus, many papers that deal with sensitive data~\cite{dou2023reducing,fabian2015privacy,volkova2015inferring} end up not sharing datasets publicly. This impedes much-needed research on these topics. 
Furthermore, creating \textit{labeled} datasets is an expensive and time-intensive task since it may require careful manual annotations \cite{smart2024discipline,yin2023annobert}; thus, reuse and sharing are important even beyond just enabling reproducibility. In several cases, the cost of creating ``gold-standard'' data can limit the scale of such datasets, which in turn limits the scale and sophistication of downstream research for understanding the problem, or machine learning models that can help prevent leakage. 

A clear and ``obvious'' solution to this dilemma is to generate a synthetic dataset that \textit{does not contain information about any particular individual, but has sufficient representation of different kinds of PII leakage seen in real datasets} so that it can be used to develop both a data-driven understanding as well as ML models that can help detect PII leakage. Achieving this requires \textit{(i)} starting with a real dataset that would be useful to the research community \textit{(ii)} compiling a list of characteristics or labels of interest, and (manually) annotating the dataset to identify different kinds of PII leakage, and \textit{(iii)} generating a synthetic dataset which looks similar to the real dataset and preserves its utility to researchers whilst being difficult to link back to the posts in the real dataset. 

Our main contribution is a novel method to create synthetically generated datasets that are equivalent to the original data in important ways, such as looking similar to and preserving the style of the original dataset, whilst still providing substantially improved privacy. Our method relies on private, locally hosted LLMs,  along with RoBERTa-based models fine-tuned for multilabel classification and span-level PII detection across 19 sensitive categories. 

Our extensive evaluation shows that the synthetic data looks similar to, and retains the key characteristics of the original dataset, so that it remains useful for researchers whilst also keeping data from real users private. We develop three metrics to demonstrate this: The first metric, \textit{reproducibility equivalence}, evaluates, the \textit{utility of the synthetic dataset to researchers} for its original intended purpose of ensuring research reproducibility.
The second metric, \textit{indistinguishability}, evaluates to what extent the two datasets look similar to \textit{humans}. Specifically, we develop a user study with participants tasked with distinguishing the synthetic posts from the original posts.
The third metric, \textit{unlinkability}, evaluates the \textit{resistance of the synthetic dataset to being deanonymized} by asking how easy it would be to find the original public social media post through a Google search with the equivalent synthetic post. We further evaluate our synthetic dataset on three metrics proposed in \cite{chim2024evaluating} for meaning, style and privacy preservation.
\vspace{-2pt} 
After validating the dataset across these criteria, we fine-tuned a RoBERTa-based span categorization model using the synthetic data, achieving strong performance in detecting PII entities across multiple categories. 

To support transparency and encourage further research, we open-source our code and make both the dataset and the modeling pipeline freely available for non-commercial research usage, at  {\url{https://netsys.surrey.ac.uk/datasets/synthetic-self-disclosure/}}. 
\section{Related Work} 
\textbf{PII-annotated datasets} {A variety of corpora now exist that label text for different categories of personally identifiable information (PII). Several focus on disclosure detection, using either binary labels (e.g., 0 for non-disclosure and 1 for disclosure), or three-way labels (e.g., no disclosure,  possible disclosure and  clear disclosure) for self-disclosures of sexual abuse \cite{chowdhury2019speak}, health conditions \cite{valizadeh2021identifying}, and personal revelations posted to a Reddit-style Korean community \cite{cho-2022}. Beyond disclosure, the MBTI9k corpus assigns one of 16 Myers-Briggs personality types to Reddit users \cite{gjurkovic2018reddit}, while the large-scale RedDust resource provides 300k Reddit posts annotated for five user attributes—profession, hobby, family status, age, and gender—for profiling tasks \cite{tigunova2020reddust}. \textit{Both these works, whilst highly useful to researchers, reveal PII about Reddit users}. A Wikipedia-based corpus labels biography sentences for five classes of personal information, but its automatic annotation procedure introduces substantial noise in the dataset \cite{hathurusinghe2021privacy}. Finally, a Kaggle competition released synthetic PII text by inserting generated names and e-mail addresses into LLM-produced passages \cite{pii-detection-removal-from-educational-data}; however, the synthetic nature of this data restricts it to a few basic PII types and fails to capture the richer, context-dependent disclosures common in social-media language.}
\par While the above classification-oriented datasets are publicly available to the research community, recent works proposing more comprehensive PII category data do not make the resultant datasets open-source due to privacy or ethical considerations. These include the eight categories of personal attributes (sex, location, marriage, age, education, occupation, place of birth, income) applied to a curated Reddit dataset~\cite{staab2024beyond}.
Although the authors claim to make a synthetic version of the original posts openly available, these need a seeded example (5 sample texts provided for the education category) to generate the posts with the provided script. Similarly, Dou et al.~\cite{dou2023reducing} propose 19 broad-coverage PII categories along `attributes' and `experiences' aspects for a PII span-annotated dataset. Our work focuses on categories specific to vulnerable people undergoing significant life transitions and extends it to include explicit as well as implicit PII mentions, together with word spans leaking the corresponding PII. 

\par \textbf{Online PII Identification Models} 
Svitlana et al. \cite {volkova2015inferring}
demonstrated a latent personal attribute prediction approach using trained log-linear models with lexical features extracted from 200 tweets per user for 5000 Twitter profiles annotated through crowdsourcing, focused around 10 demographic attributes, 5 personality traits, and three types of controlled impression behavior. Fabien et al. \cite{fabian2015privacy} 
proposed two large-scale classification models (soft-margin SVM classifiers and supervised LDA) corresponding to gender (Male/Female) and citizenship (grouped by continent), which are trained on lexical rules-driven annotated Reddit data. Considering the objective definition of privacy-sensitive content, Livio et al. \cite{bioglio2022analysis} employed advanced deep learning models to determine whether a post is sensitive. They utilized a corpus of nearly 10,000 text posts, each annotated as sensitive or non-sensitive by human evaluators. 
All of the above works do not make the associated datasets public. Zhang et al. \cite {zhang2022skillspan} introduced domain-adapted BERT models: JobBERT and JobSpan-BERT for skills and knowledge component extraction along with SKILLSPAN- a novel skill extraction dataset consisting of 14.5K sentences and over 12.5K annotated spans, which can be useful for user employment information extraction. A comprehensive study by Staab et al. \cite{staab2024beyond} analyzed the capability of LLMs such as GPT 3.5, Palm 2 Text, Llama-2 family, etc., to infer PIIs about Reddit post authors from the post text, concluding that LLMs have human-like performance in detecting various PIIs.  Earlier works have proposed various disclosure detection models by modelling it as a binary~\cite{chowdhury2019speak} or multi-label~\cite{balani2015detecting, valizadeh2021identifying,cho-2022} text classification task, for the presence, absence or likelihood of self-disclosure at sentence level \cite{valizadeh2021identifying, chowdhury2019speak, cho-2022}. A more relevant approach is proposed by Duo et al.~\cite{dou2023reducing} who fine-tune a RoBERTa-large model to identify text spans of self-disclosure along 19 proposed categories of PII. 
\vspace{-8pt}
\section{Data Collection}
\label{sec:data_collection}
To identify vulnerable communities on Reddit, we analyzed the top 1,000 largest subreddits listed at \url{https://www.reddit.com/best/communities/1/}, focusing on those that function as support networks for populations aligning with the definition of vulnerability outlined in Section \ref{sec:introduction}.
Notably, r/ADHD (ranked 445th, with \~2 million members) and r/lgbt (ranked 694th, with \~1.2 million members) emerged as two of the most prominent subreddits serving as support communities for vulnerable populations. For this study, we focus on the members of r/lgbt due to its explicit engagement with issues of identity, discrimination, and community support -- factors closely aligned with PII leakage risks. We crawled posts from 2016 to 2020 from r/lgbt using the PushShift API \cite{pushshift}. We deliberately rely on old posts to avoid potential risks from leakage that we may not have anticipated. We identified the 500 most active \texttt{r/lgbt} subreddit members, and then collated posts and comments they may have made across different subreddits.
This enables us to curate a dataset with longitudinal posts from active users, allowing identification of incremental self-disclosures. Among these 500 users, 100 have deactivated their accounts, resulting in a collection of 401,983 records from 400 users. A total of 8679 posts and comments from NSFW (Not Safe For Work) or Over\_18 labelled subreddits are filtered out. 

We streamline our approach to sift through user-generated content, specifically honing in on statements where individuals discuss themselves or their collective experiences. This strategy emphasizes focusing on retaining posts and comments that include first-person references, containing pronouns and words such as `I,' `me,' `myself,' `my,' `mine,' `we,' `us,' `our,' and `ours.' We refine our dataset by utilizing regular expressions to recognize these linguistic markers. Following this, posts and comments containing less than three words are removed, resulting in 65,282 records. Finally, 5\% of the remaining posts (i.e., 3264 posts) are randomly sampled for data annotation. These posts originate from 293 distinct subreddits, introducing a diverse range of content. These thematically distinct communities helped to include varied vocabularies and discourse styles in the dataset, thereby strengthening the ability to generalise to unseen subreddits and even other social-media platforms. We augment each post with the name of the subreddit where it was posted, as they serve as markers for potential PII-revealing information. For instance, location-based and community-specific subreddits can provide key insights into user identities.
\begin{table*}[t]
\caption{PII taxonomy with related statistics and example posts}
\label{tab:PII-stat}
\resizebox{\textwidth}{!}{
\begin{tabular}{llll}
\hline
\textbf{PII} & \textbf{\begin{tabular}[c]{@{}l@{}}Number \\ of spans\end{tabular}} & 
\textbf{\begin{tabular}[c]{@{}l@{}}Average \\ span length\end{tabular}}
& \textbf{Example}            \\ \hline
\begin{tabular}[c]{@{}l@{}}Name \\ Birthdate\\ Location\\ Country\\ Marital Status\\ Religion\\ Ethnicity/Race\\ Gender\\ Parenthood\\ Age\\ Sexuality\\ Medical Information\\ Employment 
\\Relationship\\Family\\Gender-Age\\Mental Health \\Physical Appearance \\Degree/Designation
       \end{tabular} & \begin{tabular}[c]{@{}l@{}}118\\ 19\\ 173\\86\\57\\59\\90\\610\\54\\165\\462\\130\\288 \\91\\269\\29\\200\\44\\12\end{tabular}              & \begin{tabular}[c]{@{}l@{}} 17.18\\ 26.68\\ 34.84\\ 26.57\\13.46\\28.49\\14.92\\23.33\\12.59\\12.93\\22.01\\32.75 \\27.61 \\15.95 \\11.49 \\8.72 \\28.18 \\21.52 \\33.75  \end{tabular} & \begin{tabular}[c]{@{}l@{}}Hey fellow Redditors! \textcolor{red}{I'm k**t**n}.\\ \textcolor{red}{I'm turning 18 in 3 days} and I am feeling lost.\\ \textcolor{red}{I was walking in downtown Brooklyn very close to the Brooklyn Bridge}.\\ \textcolor{red}{I'm from the Philippines} and I'm excited for my new journey.\\ \textcolor{red}{My wife}, not me, got a text message.\\ \textcolor{red}{I was raised Catholic} and I have seen things on the Atheism.\\ \textcolor{red}{I'm a British person} and I think I might have found my long-lost grandpa.\\ I don't tell people in my day-to-day life that \textcolor{red}{I'm transgender}.\\ \textcolor{red}{My son} is a huge Mustang fan.\\ I'm a 14 year old and life's been pretty tough for me lately.\\ Isn't it so damn lovely! Left \textcolor{red}{my lil lesbian heart} all warm \& fuzzy.\\ \textcolor{red}{I've been getting chemo and radiation to the abdomen}\\ \textcolor{red}{I'm a medical intern} in I**on*s*a.\\\textcolor{red}{I'm dating a binary woman}\\ \textcolor{red}{My dad} passed away when I was 10 due to a heart attack.\\ Hey \textcolor{red}{21M here}, pm me if you're down to play some online games.\\ \textcolor{red}{I've had ADHD} all my life, but I've only been recently diagnosed.\\
      Help a bro out! \textcolor{red}{I'm a 5'9" tall, 180 lbs}.\\ \textcolor{red}{As a CS major, I'm} used to running multiple programs simultaneously.
       \\
       
       \end{tabular} \\ \hline
\end{tabular}}
\end{table*}
\subsection{Data annotation}
Two annotators with domain expertise started the annotation process as crowdsourcing often leads to lower quality annotations \cite{dou2023reducing}. Annotation guidelines were formulated after reviewing 500 records with posts and comments. We used an open-source annotation tool, Doccano \cite{doccano}, that provides a user-friendly platform. Annotators marked personal information disclosing text spans. These spans include PII along with self-referential text to preserve context. For instance, instead of highlighting just "As a transgender," we highlight "As a transgender, I." 
 After annotation, 1,183 of 3,264 posts were found to contain PII-revealing text, while 2,081 did not. Together, these posts form our gold-standard annotated dataset.
Inter-annotator agreement (IAA) metrics were used to improve the annotation guidelines and ensure good-quality annotated data. Commonly used IAA measures such as Cohen’s Kappa 
and Fleiss’ Kappa 
require the precise definition of negative samples and hence are not suggested for the span-based annotations. The recommended metric for span-based annotations is the pairwise F1 score \cite{hripcsak2005agreement}. We considered two annotations to agree if they had any overlapping words (partial span) and the same label. The pairwise F1 Score with overlap is 0.8275 for inter-annotator agreement. On average, the overlapping portion of the agreed-upon spans was 70.27\%. Table~\ref{tab:PII-stat} shows the 19 categories annotated for, including illustrative examples.
\section{Synthetic Data Generation}
\subsection{Text Generation Models}
Synthetic data generation used three LLMs -- Llama 2-7B, Llama 3-8B, and Zephyr. This process involved a 1:3 mapping, where a single original post served as the input source for three synthetic posts: one by each LLM.
We randomly selected 50 posts and generated synthetic versions using temperatures ranging from 0.5 to 1 to determine the optimal temperature settings for the LLMs. For each LLM, we selected the temperature that produced the lowest average cosine similarity between the original and generated posts, ensuring the generated posts were sufficiently distinct and non-linkable to the originals. For Llama 2-7B,  we used the Llama 2-Chat model which is fine-tuned for dialogue generation. We use instruction prompting with a maximum sequence length of 1024 tokens and a batch size of 8. During generation, a temperature of 1 and nucleus sampling with top\_p of 0.9 were employed to balance diversity and coherence in the outputs.
We used the  zephyr-7b-beta model,  a fine-tuned version of Mistral-7B \cite{tunstall2024zephyr}, provided by HuggingFace API with temperature=1 and top\_p=0.95 for generating the synthetic data. For Llama 3-8B, we used the Meta-Llama-3-8B-Instruct model with temperature=0.9 and top\_p=0.9. For Llama3 and Zephyr, we used default sampling parameter settings to ensure adequate synthetic data quality.
\begin{figure}[!t]
\centering
\fbox{
\parbox{\textwidth}{
\textbf{Prompt 1:} 
 Change the original post following these rules:
\begin{enumerate}
  \item  Replace all non-sensitive private information such as age, dob, religion, gender, marital status, race, ethnicity, employment, location, sexuality, and parenthood with other non-sensitive private information that retains the context. Replace the organization name with any other organization that serves the same purpose without generalization.
\item Change specific codes, IDs, numbers, and names with different codes, IDs, numbers, and names, respectively.
\item Generate a post that matches the same style and tone as the original post. If the original post contains spelling errors, strong language, or informal expressions, ensure that the synthetic post reflects the same characteristics.
\item Use common internet abbreviations, slang, emoticons, and expressions where appropriate, keeping the overall feel and context of the original post intact.
\item  Don't give the title of the post.
\end{enumerate}
\textbf{Prompt 2:} 
The first line of the original text tells about the subreddit name in which the original post was posted. Change the name of the subreddit to another subreddit of a similar kind. \\
\textbf{System Prompt:} 
You are a story recreator who takes the information from the original post,
and then makes a different story with similar kinds of personal information. You want to minimize the chance of finding the link between the stories. Generate the post following this format:\\
"Changed Post":
}
}
\caption{System and instruction prompts}
\label{fig:prompt}
\end{figure}

 Instruction tuning enables LLMs to follow user instructions and perform zero-shot generalization \cite{hu2024finetuning}. We incorporated multiple instructions in the prompt to generate synthetic data. The prompt provided to the model, as shown in Figure \ref{fig:prompt}, consists of two instruction prompts and a system prompt. This approach was adopted because Llama models often struggle to follow a sequence of instructions within a single query, occasionally ignoring or misinterpreting parts of the instructions. 
Consequently, we employed sequential instruction tuning, using a two-step process: first, to modify the content, and second, to change the subreddit name.
Given the input $x$,  suppose $p_1$ is the prompt for the first step of the task and $p_2 $ is the prompt for the second step. The final output is obtained as $ \hat{y_2} \sim p(y_2 |p_2, \hat{y_1}; \theta_{LLM})$ and $\hat{y_1} \sim p(y_1 |p_1, x; \theta_{LLM})$.
The system prompt contains initial instructions sent to the API that define the behavior of the models and guide their response generation. 
\begin{table}[!t]
\caption{Summary statistics of generated dataset variants using LLMs}
\label{tab:sdataset}
\begin{tabular}{@{}lllll@{}}
\toprule
\textbf{Dataset}                                                                                 & \textbf{\begin{tabular}[c]{@{}l@{}}Number of\\    rounds\end{tabular}} & \textbf{\begin{tabular}[c]{@{}l@{}}Dataset\\    Size\end{tabular}} & \textbf{\begin{tabular}[c]{@{}l@{}}Size after non-linkability\\    threshold\end{tabular}} & \textbf{\begin{tabular}[c]{@{}l@{}}Number of\\    Spans\end{tabular}} \\ \midrule
\begin{tabular}[c]{@{}l@{}}Llama2-genearted \\ Llama3-genearted \\ Zephyr-genearted\end{tabular} & \begin{tabular}[c]{@{}l@{}}3\\    3\\    1\end{tabular}                & \begin{tabular}[c]{@{}l@{}}971\\   913\\   1054\end{tabular}       & \begin{tabular}[c]{@{}l@{}}954\\   900\\   1034\end{tabular}                                & \begin{tabular}[c]{@{}l@{}}1660\\   1559\\   1919\end{tabular}        \\ \bottomrule
\end{tabular}
\end{table}
 We are using real Reddit posts as seeds to generate synthetic content that might have text describing sensitive issues, such as transphobic activities or mental health issues. 
Following ethical AI governance and content moderation, Llama models sometimes refuse to generate synthetic content around this sensitive topic. Additionally, requests to change PII raise concerns about potential privacy violations, prompting the AI to refuse these alterations to prevent misuse. 
 Therefore, we did two additional rounds of synthetic posts generation for the posts for which synthetic data generation was denied. 
 We did not observe this behavior in the case of Zephyr and obtained synthetic data without the need for additional rounds.
After generation, the synthetic data was preprocessed and manually annotated for PII-leaking spans.  Table \ref{tab:sdataset} presents the synthetic data description. We have different counts of generated posts for each LLM as Llama-based models refused to generate for some of the posts even after the third text-generation round. Further, some generated posts that do not contain any PII are discarded after annotation. 
\vspace{-10pt}
\section{Data quality Evaluation}
We evaluate the synthetically generated dataset by developing three metrics that ensure the quality of generated text, in terms of being usable as a stand-in for the original dataset and ``looking similar'' to it, while at the same time being difficult to reverse engineer the original text given a synthetic post: \textit{1)} Reproducibility equivalence, \textit{2)} Indistinguishability, \textit{3)} Unlinkability. 
\subsection{Reproducibility equivalence}
This metric ensures that synthetically generated data must preserve the same utility value as the original data.  Our goal is to train a model on the synthetic dataset with a similar or better performance as when trained on the original data. To check this, we utilized the Hugging Face TrainingArguments class to fine-tune a pre-trained RoBERTa model \cite{liu2019roberta} for a multi-label classification task of predicting which (if any) of the 19 PII categories (Table~\ref{tab:PII-stat}) a post contains. Specifically, we set the number of training epochs to 50 with default learning rate and weight\_decay=0.01. The batch size was kept at 8. Datasets are split into an 80-20\% split for training and testing sets. The Table ~\ref{tab:classification} demonstrates that the performance metrics of the multilabel classifier on synthetic datasets are comparable to those on the original dataset, demonstrating that the synthetically generated datasets have the same utility as the original dataset.
We also fine-tuned a custom RoBERTa-based span categorization model, designed as a multilabel token classification task. Training was performed using the \textit{BCEWithLogitsLoss} function, which combines sigmoid activation with binary cross-entropy loss applied at the token level. The model was trained entirely on synthetic data generated by the three different LLMs, using an 80-20\%  train-test split. The model demonstrated solid performance, achieving a token-level macro F1 score of 0.6965 and a partial span-level F1 score of 0.70 (with a minimum 50\% overlap). These results are notable given the large number of target categories and surpass the performance reported in \cite{dou2023reducing}, which tackles a comparable set of PII classification categories.
\begin{figure}[!t]
 \centering
  \includegraphics[width=0.5\textwidth] {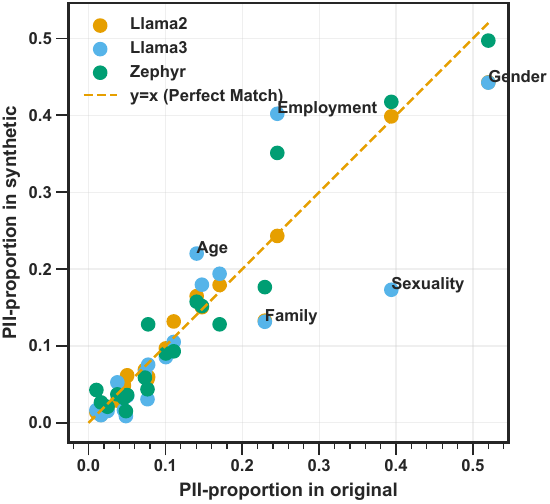}
  \caption{ PII-proportion in original and synthetic data}
  \label{fig:scatter}
\end{figure}

A second aspect of reproducibility is whether the synthetic dataset has a similar mixture of different PII categories as the original. This may be required when using such a dataset in some applications, although in a synthetically generated dataset, it would be entirely possible to selectively boost or suppress the prevalence of PII categories if the application requires this. Figure~\ref{fig:scatter} compares the proportion of different PII categories in synthetic datasets generated by Llama2, Llama3, and Zephyr against the original dataset. The red dashed line represents an ideal 1:1 match, where synthetic data would perfectly preserve PII proportions from the original dataset. 
One notable observation is that Llama2 tends to preserve the original proportion of PII proportions more faithfully. This suggests that Llama2 is less creative and more conservative in its text generation, leading to a distribution that remains close to the original. In contrast, Llama3 demonstrates greater creativity, possibly paraphrasing or altering details in a way that introduces bias towards certain PII categories. In particular, the Llama3-generated synthetic dataset has fewer gender, sexuality, and family-related PII details than the original, but over-expresses age-related PII disclosure posts. Zephyr, which is designed for more controlled generation and instruction-following, shows a moderate deviation -- less extreme than Llama3 but still differing from the original data in key categories like Employment and Family. 
Zephyr, being more instruction-optimized, appears to balance preserving PII distribution while still introducing some variation, making it an intermediate case between the two. This highlights the challenge of balancing synthetic data fidelity and text diversity, particularly in privacy-sensitive applications.



\begin{table}[!t]
\centering
\caption{Roberta-based multi-label classifier performance on different datasets}
\label{tab:classification}
\begin{tabular}{@{}lllll@{}}
\toprule
\textbf{Dataset} & \textbf{Accuracy} & \textbf{Precision} & \textbf{Recall} & \textbf{F1- score} \\ \midrule
\textbf{Original}                      & 0.6772                  & 0.8311                   & 0.8096                &0.85374                   \\
\textbf{Llama2 generated}                        & 0.6359            & 0.8407             & 0.8507          & 0.8457            \\
\textbf{Llama3 generated}                        & 0.6448            & 0.8369             & 0.8690          & 0.8527              \\
\textbf{Zephyr generated}                        &0.6462                   & 0.8666                   & 0.8471                &  0.8871                 \\ \bottomrule
\end{tabular}
\end{table}

\subsection{Indistinguishability}
\begin{table}[b]
\centering
\caption{Distinguishing probability of LLM-generated text}
\label{tab:probability}
\begin{tabular}{@{}llll@{}}
\toprule
                              & \textbf{Set 1} & \textbf{Set 2} & \textbf{Set 3} \\ \midrule
\textbf{Observed Probability} & 0.54           & 0.34           & 0.30           \\
\textbf{Expected Probability} & 0.5            & 0.33           & 0.25           \\ \bottomrule
\end{tabular}
\end{table}

\begin{table*}[!t]
  \centering
\caption{Category of reasons provided by survey participants}
\label{tab:reason}
\resizebox{\textwidth}{!}{
\begin{tabular}{@{}ll@{}}
\toprule
\textbf{Categories}   & \textbf{Keywords}                                                                                                                                                                                                \\ \midrule
Content Detail         & Short, Very brief, Over-elaboration, Condensed Paragraph, Detailed                                                                                                                                               \\
Tone and Style            & Natural, How it sounds, Genuine, CPU stating facts, Formal or  Formulatic Language                                                                                                         \\
Language Characteristics & \begin{tabular}[c]{@{}l@{}}Offensive and strong language, Personal language, Swear words, Emotive Language, \\ Repetitive and sterile way, Slang used, Use of acronym, Abbreviations\end{tabular} \\
Grammar and Structure               & No grammar, Poor sentence structure, Spelling error, Confusion of phraseology                                                                                                                    \\ \bottomrule
\end{tabular}}
\end{table*}
The previous section showed that the synthetic dataset can serve as a drop-in replacement for the original when training automated models. Next, we ask whether synthetic data appears similar to the original Reddit posts for humans, i.e., it should be hard for humans to distinguish between the original and synthetic posts if they are not told which is which. To evaluate this, we recruited 100 participants through Prolific, an online research platform to recruit people for participation in a research study \cite{palan2018prolific}. Participants were given three sets of identical tasks in order to distinguish which text/post appears to be written by a human rather than generated by any Large Language Model: Set 1 consists of 2 posts i.e., one post is an original text written by a human, the other is generated by an LLM; a random guess would have 50\% chance of success). Set 2 consists of 3 posts i.e., one original post and 2 LLM-generated posts; a random guess would have 33.33\% chance of success. Set 3 consists of 4 posts i.e., one original post and 3 LLM-generated posts; a random guess would have 25\% chance of success. The posts in each set were randomly selected from both synthetic and original posts. To discourage the Prolific participants from randomly guessing an answer, they were asked to also answer a follow-up question aimed at eliciting the factors they used to make their choices.   
We computed the chi-square deviation to determine how much the observed probability distribution of selecting the correct original post differs from the expected probability distribution, i.e., the probability of random selection as shown in Table \ref{tab:probability}. 
The null hypothesis ($H_0$) posits that participants cannot reliably distinguish human-written posts from LLM-generated posts (i.e., no difference between observed and expected distributions), while the alternative hypothesis ($H_1$) suggests that participants are able to make this distinction. The analysis yielded a chi-square statistic of 1.35 and a corresponding \textit{p}-value of 0.51, indicating no statistically significant deviation from random selection under the null hypothesis. 
\par Thus, we fail to reject the null hypothesis. We achieve our indistinguishability requirement, i.e., our study participants \textit{are not able to distinguish between human-written and LLM-generated text}. 
Some participants were able to recognise the original one, for instance, one noted: ``grammar mistakes and poor flow,'' in the human-generated text, while another described the original post as ``erratic.'' The reasons are mainly classified into four categories as shown in Table \ref{tab:reason}. Yet, even those who selected correctly expressed difficulty, e.g., one said: \textit{``I think this one is the most believable [as human generated] but all seem so human sounding it's quite hard to choose''}. 

\subsection{Unlinkability}
\begin{figure}[!b]
  \centering
  \includegraphics[width=0.6\linewidth]{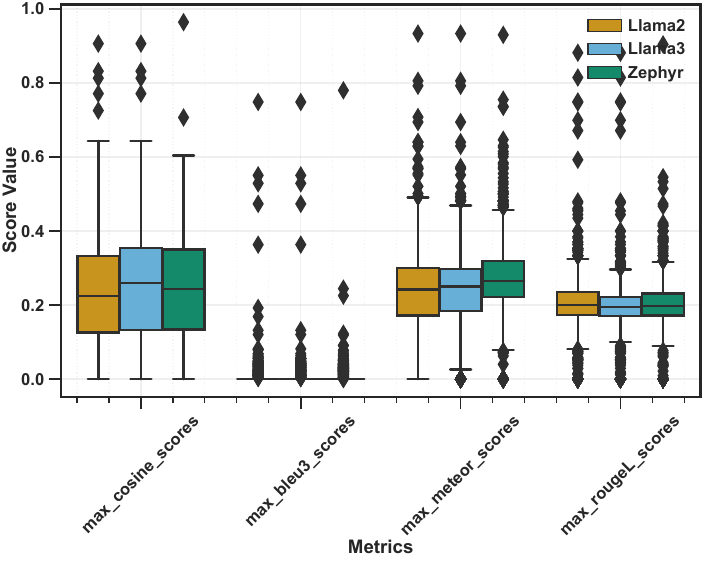}
  \caption{ Text similarity metrics between synthetic posts and top=10 Google search results}
  \label{fig:metrics}
\end{figure}
A primary motive behind synthetic data generation is that publishing original data, highlighting users' self-disclosure texts, can create significant privacy risks. Therefore, this metric ensures that synthetic posts should not lead to the original posts they intend to mimic. To test this, we search for the text of the synthetic post using the Google search API and extract the top-10 search results. We aim to find the original post this way. We limit our results to the top 10 because the first page of Google Search typically contains no more than 10 entries. However, we performed a sensitivity analysis and checked that similar results hold even if we go to the third page of Google search results, and on average, beyond the 6th result, we begin encountering non-Reddit links.  

We ignore all search results that do not lead directly to Reddit. To check that the Reddit link that matches a synthetic post is not related to the original post it is generated from,  we calculate similarity scores between the synthetic text and posts/comments present on the Reddit link. We used cosine similarity score, Bleu score, Meteor, and Rouge to quantify how similar the content is \cite{liu2023g}. Using them together provides a more comprehensive evaluation.  
The Figure \ref{fig:metrics} demonstrates a strong balance between linguistic diversity and semantic fidelity. The synthetic posts of LLMs demonstrate significant rephrasing and structural variation, as reflected by the very low \textit{max\_blue3\_scores} values. 
The generated posts have significantly different wording, phrasing, or structure from the original posts. This is due to the LLMs prioritizing diversity or rephrasing text. 
Low content similarity scores for all the metrics establish that synthetic data is non-linkable to original data. 
Further, to ensure the anonymity of the synthetic dataset, we discarded synthetic posts with a Meteor score greater than 0.5 to avoid any chance of tracing the synthetic post back to the original post.
\vspace{-10pt}  
\subsection{Supplementary Evaluation }
To complement the above analysis, we also test for three core aspects of data quality expected in synthetically generated texts, as identified by Chim et al. \cite{chim2024evaluating} recently. These metrics, which are independent of the data generation strategy, include \textit{BERTScore} for meaning preservation, \textit{style embedding similarity} for style preservation, and \textit{divergence} as a proxy for privacy. BERTScore goes beyond exact word matches (relied upon by BLEU, ROUGE etc.), and aims to more fully capture meaning preservation. 
Controlling stylistic elements and ensuring linguistic diversity matter, as variation should promote generalization without compromising label validity. We extract idiolect embeddings using pooled RoBERTa representations for both original and synthetic posts and compare their writing patterns like sentence structure, tone, and phrasing.
Divergence is calculated by measuring the BLEU score between the source and synthetic text, with the divergence defined as $1 - BLEU(s, t)$. This approach is effective for privacy metrics, as it quantifies the surface-form dissimilarity, serving as a proxy for verbatim memorization. The table \ref{tab:tacl} shows that Zephyr generally performs best. The prominent difference is in terms of \textit{divergence}.  
\begin{table}[t]
  \centering
\caption{Synthetic data evaluation of metrics from \cite{chim2024evaluating} }
\label{tab:tacl}
\begin{tabular}{@{}llll@{}}
\toprule
\textbf{Metrics}                                                                                    & \textbf{Llama2} & \textbf{Llama3} & \textbf{Zephyr} \\ \midrule  
\textbf{Meaning preservation (BERTScore)}
                & 0.92            & 0.87            & 0.90            \\
\textbf{Style Preservation (Style similarity)} & 0.98            & 0.97            & 0.96            \\
\textbf{Privacy preservation (Divergence)}                & 0.58            & 0.89            & 0.95            \\ \bottomrule
\end{tabular}
\end{table}
\section{Conclusion}
In this paper, we introduced a taxonomy of 19 PII-revealing categories relevant to vulnerable populations and generated a synthetic PII-labeled span dataset using the capabilities of LLMs, specifically LLaMA2-7B, LLaMA3-8B, and Zephyr-7B. This dataset addresses the scarcity of labeled, privacy-preserving data and enables reproducible research into machine learning models for detecting PII-revealing content. We demonstrated the practical utility of this dataset by fine-tuning RoBERTa-based models for multilabel classification and token-level span categorization tasks. Despite the inherent complexity and breadth of the 19 PII categories, our span-level model achieved strong performance (macro F1 = 0.70) when trained entirely on synthetic data, showcasing its effectiveness for real-world applications. We have begun integrating this work into tools such as the InsightWatcher browser plugin \cite{haq2025exploringselfdisclosure}, which detects self-disclosures on social media platforms. To support further research, we release our comprehensive PII-labeled synthetic dataset along with our codebase, enabling others to generate custom synthetic datasets and adapt our methods to their specific needs. In our future work, we will examine how this can be done in a principled manner, for example, in order to overcome biases which may exist in the original datasets. We also plan to extend the dataset to better represent categories with lower occurrence in our original dataset, such as Degree/Designation and Physical Appearance.

We also hope that our method of generating synthetic equivalents of datasets is generalisable and will lead to better reproducibility of research that relies on data otherwise difficult to share due to concerns such as preserving the privacy of participants. In particular, our three metrics of evaluation --- reproducibility equivalence, indistinguishability and unlinkability --- can help quantify the utility and privacy preservation of synthetic data in other contexts, thereby helping researchers better balance between these conflicting goals. 

\bibliographystyle{splncs03}
\bibliography{synthetic}

\end{document}